\documentclass[conference,a4paper]{IEEEtran}

\usepackage{epsfig}
\usepackage{subfigure}
\usepackage{amsmath}
\usepackage{amssymb}
\usepackage{amsthm}
\usepackage{graphicx}
\usepackage{subfigure}
\usepackage{multirow}
\usepackage{flushend}
\usepackage{url}
\usepackage{color}
\newcommand{\bega}{\begin{eqnarray}}
\newcommand{\ega}{\end{eqnarray}}
\newcommand{\bb}{\begin{equation}}
\newcommand{\ee}{\end{equation}}

\newtheorem{thm}{Theorem}[section]
\newtheorem{theorem}[thm]{Theorem}
\newtheorem{corollary}[thm]{Corollary}

\newtheorem{conjecture}[thm]{Conjecture}

\title{Belief Propagation for Linear Programming}
\author{
\IEEEauthorblockN{Andrew E. Gelfand}
\IEEEauthorblockA{Department of Computer Science\\
University of California, Irvine\\
Irvine, CA  92697-3435\\
Email: agelfand@ics.uci.edu}
\and
\IEEEauthorblockN{Jinwoo Shin}
\IEEEauthorblockA{
Mathematical Sciences Department\\
IBM T. J. Watson Research,\\
Yorktown Heights, NY, USA 10598\\
Email: jshin@us.ibm.com}
\and
\IEEEauthorblockN{Michael Chertkov}
\IEEEauthorblockA{Theoretical Divison\\
Los Alamos National Lab\\
Los Alamos, NM 87545, USA\\
Email: chertkov@lanl.gov}
}

\begin{document}
\maketitle
\begin{abstract}
Belief Propagation (BP) is a popular, distributed heuristic for performing MAP computations in Graphical Models. BP can be interpreted, from a variational perspective, as minimizing the Bethe Free Energy (BFE). BP can also be used to solve a special class of Linear Programming (LP) problems. For this class of problems, MAP inference can be stated as an integer LP with an LP relaxation that coincides with minimization of the BFE at ``zero temperature". We generalize these prior results and establish a tight characterization of the LP problems that can be formulated as an equivalent LP relaxation of MAP inference. Moreover, we suggest an efficient, iterative annealing BP algorithm for solving this broader class of LP problems. We demonstrate the algorithm's performance on a set of weighted matching problems by using it as a cutting plane method to solve a sequence of LPs tightened by adding ``blossom'' inequalities.
\end{abstract}

\section{Introduction}

Graphical Models (GMs) provide a useful representation for reasoning in a range of scientific fields \cite{05YFW,08RU,09MM,08WJ}. Such models use a graph structure to encode a joint probability distribution, where vertices correspond to random variables and edges (or lack thereof) specify conditional independencies.

An important inference task in many applications involving GMs is finding the most likely assignment to the variables in a GM - the Maximum-A-Posteriori (MAP) configuration. Belief Propagation (BP) is a much celebrated algorithm for approximately solving the MAP inference problem. BP is an iterative, message passing algorithm that is exact on tree structured GMs, but has empirically been shown to give good results even on GMs with loops. Its main appeal is that it is naturally suited for a distributed implementation.

It was recently shown that BP is exact for a certain class of GMs with loops. This inspiring result was shown for GMs in which well known optimization problems - namely, the matching problem \cite{08BSS,11SMW} and min-cost network flow problem \cite{10GSY} - were posed as MAP inference tasks. In the weighted matching case, the original combinatorial optimization problem can can be expressed as a binary Integer Linear Program (ILP). In certain cases (e.g. in bi-partite graphs), solving the LP relaxation to the matching ILP yields an integral solution.

The MAP inference task can also be formulated as an ILP in GMs with discrete variables. The LP relaxation to the MAP ILP, which we refer to herein as BPLP, arises by relaxing the integrality constraint on the discrete variables. When the weighted matching problem is formulated as a MAP inference task, BPLP is equivalent to the relaxed matching ILP - explaining the success of BP in these GMs \cite{08BSS,11SMW,08Che}! The connection between LP relaxations and BPLP (also called LP-decoding) has also been discussed in the coding literature \cite{05FWK,06VK,08CS,06TS,09DGW}.

This line of work established a solid theoretical link between message passing algorithms and optimization theory. It provides a practical certificate of exactness/integrality for MAP inference when using BP and has also suggested strategies for improving upon BP's results, by adding constraints that reduce the BPLP integrality gap \cite{08Joh,10Son,11KJC}.

Motivated by this prior work, our manuscript characterizes the class of binary ILPs for which LP=BPLP, i.e. where the LP relaxation of an ILP is equivalent to BPLP, the LP relaxation of the MAP formulation of the problem.  While standard BP is an approximation algorithm that is not guaranteed to converge to a correct answer for the LP, we provide an annealing version of BP that converges to the correct answer as long as LP=BPLP. Establishing this relationship allows us to use BP (or its variants) to efficiently approximate MAP inference in the special class of binary ILPs. We extend the work in \cite{08BSS,11SMW} by empirically demonstrating that annealing BP can be used to solve LP-relaxations to the weighted matching problems requiring Edmonds' blossom inequalities. In particular, we use annealing BP to solve a sequence of successively tightened LP relaxations. If coupled with the method for finding a tight LP relaxation of `polynomial' size in \cite{12CVV}, annealing BP could be used in a novel, distributed approximation algorithm for the weighted matching problem.

The material in the manuscript is organized as follows.
Section \ref{sec:pre} introduces GMs, BP and the class of LPs of interest.
Section \ref{sec:main} provides our main result. Sections \ref{sec:annealbp} and \ref{sec:exp} describe our annealed BP algorithm and demonstrate its utility as an LP-solver.

\vspace{-2pt}
\section{Preliminaries}\label{sec:pre}
\vspace{-6pt}

\subsection{Graphical Model}
\vspace{-2pt}
Let $Z=[Z_i]$ be a collection of $n$ random variables, each of which takes values in a finite alphabet $Z_i = z_i \in \Omega$. Let the joint probability distribution of $Z \in \Omega^n$ factor into a product of real-valued, positive functions $\{\psi_{\alpha} : \alpha \in F\}$ each defined over a subset of the variables:
$$\Pr[Z=z]~\propto~\prod_{\alpha\in F} \psi_{\alpha} (z_\alpha),$$
where $z_{\alpha} = [ z_i : i \in \alpha ]$ are the arguments of factor $\alpha$.
$z$ is called a valid assignment if $\Pr[Z=z]>0$. The MAP assignment $z^*$ is defined as:
\begin{equation}
z^*~=~\arg\max_{z\in \Omega^{n}} \Pr[Z=z].
\label{MAP}
\end{equation}
A Graphical Model (GM) represents the above factorization using a bi-partite graph, known as a \emph{Factor Graph} \cite{01KFL}, where each factor $\alpha \in F$ is connected to the variables in its argument. (See Figure \ref{fig:LP_FactorGraph} for an example).

\vspace{-2pt}
\subsection{Integer Linear Programming as MAP}\label{sec:ILP_as_MAP}

Consider the following ILP (Integer Linear Program):
\begin{align}
\mbox{ILP}:\qquad&\max~ c\cdot x\notag\\
&\mbox{s.t.}\qquad Ax\leq d,\qquad x_i \in \{0,1\}
\label{ILP}
\end{align}
where $i=1,\cdots,n$, $j=1,\cdots,k$, $c=[c_i],d=[d_j]$ are integer (column) vectors and $A=[A_{ji}]$
is an integer matrix.

\begin{figure}[t!]
\begin{center}
\includegraphics[width=6cm]{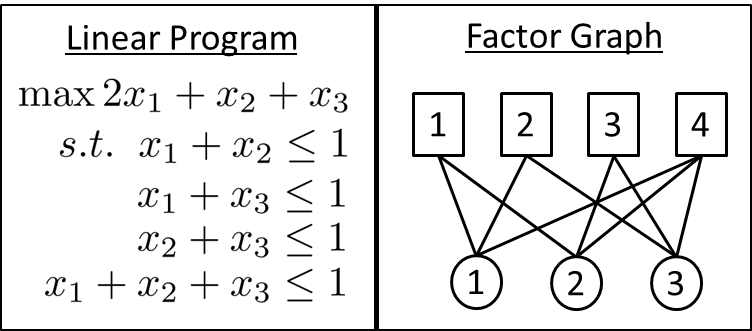}
\end{center}
\vspace{-6pt}
\caption{\small Example of a Factor Graph corresponding to a Linear Program. Factor nodes are depicted by squares and variable nodes by circles. Each factor node corresponds to one of the $4$ inequalities.}
\label{fig:LP_FactorGraph}
\vspace{-8pt}
\end{figure}

The ILP in (\ref{ILP}) can be formulated as a MAP inference task by constructing a suitable binary GM. Let $X=[X_i]\in \{0,1\}^n$ be a set of binary random variables associated with each variable in (\ref{ILP}) and consider the probability distribution:
\begin{eqnarray}
&&\Pr[X=x] ~\propto~ \prod_{i}e^{c_i x_i} \prod_{j} \psi_j(x_{_j}), \label{GM-ILP}\\
&& \psi_j(x_{\mathcal S_j})=\begin{cases}
1, &\mbox{if}~ (A x)_j\leq d_j\\
0,&\mbox{otherwise}
\end{cases},
\label{psi-ILP}
\end{eqnarray}
where every row of matrix $A$ is associated with a factor $\psi_j$ defined over a subset of the variables $x_{\mathcal S_j}$, where $\mathcal S_j = \{i : A_{ji} \neq 0 \}$. It is clear that $\Pr[X=x] \propto \prod_{i}e^{c_i x_i}$ for any `feasible' assignment satisfying the linear constraints in (\ref{ILP}) and $\Pr[X=x] = 0$ otherwise. An illustration of this transformation is shown in Figure \ref{fig:LP_FactorGraph}.

The LP relaxation of (\ref{ILP}) replaces the integrality constraints with inequalities:
\begin{align}
\mbox{LP}:\qquad&\max~ c\cdot x\notag\\
&\mbox{s.t.}\qquad Ax\leq d,\qquad0\leq x_i\leq 1.
\label{LP}
\end{align}

The LP relaxation of the ILP optimizes over a larger polytope. We will use the notation $A \leq B$ to indicate that every feasible point of polytope $A$ is a feasible point of polytope $B$ and $A \geq B$ to indicate the converse. Note that while ILP $\leq$ LP is true, ILP $\geq$ LP is \emph{not} true in general.

\vspace{-2pt}
\subsection{Bethe Free Energy and BPLP}
\vspace{-2pt}

Belief Propagation (BP) is an algorithm for approximately computing marginals that works by sending messages along the edges of the factor graph. We describe the algorithm for the GM in (\ref{GM-ILP}). Messages from factor node $j$ to variable node $i$ are denoted $m_{ji}$ and messages in the opposite direction are denoted $m_{ij}$. The messages are updated as follows: for each $x_i\in\{0,1\}$,
\begin{eqnarray}
m_{ji}(x_i) & \leftarrow & \sum_{x_{\mathcal S_j} \setminus x_i} \psi_j(x_{\mathcal S_j})^{1/T} \prod_{k \in \mathcal S_j \setminus i} m_{kj}(x_k) \nonumber \\
m_{ij}(x_i) & \leftarrow & \exp\left(\frac{c_i}{T} x_i \right) \prod_{k \in \mathcal E_i \setminus j} m_{ki}(x_i) \nonumber
\end{eqnarray}
where we have introduced a parameter $T > 0$ (called temperature) and $\mathcal E_i = \{j : A_{ji} \neq 0 \}$.

Each factor or variable node in the factor graph is associated with a belief $b_j(x_{\mathcal S_j})$ and $b_i(x_i)$, respectively. The beliefs are calculated from the messages as:
\begin{eqnarray}
b_j(x_{\mathcal S_j}) & \propto & \psi_j(x_{\mathcal S_j})^{1/T} \prod_{k \in \mathcal S_j} m_{kj}(x_k) \nonumber \\
b_i(x_i) & \propto & \exp\left(\frac{c_i}{T} x_i \right) \prod_{k \in \mathcal E_i} m_{ki}(x_i), \nonumber
\end{eqnarray}
where $\sum_{x_i} b_i (x_i) = 1$ and $\sum_{x_{\mathcal S_j}} b_j (x_{\mathcal S_j}) = 1$.

BP for the GM in (\ref{GM-ILP}) can be interpreted as a variational optimization procedure in which the messages and beliefs minimize the Bethe Free Energy (BFE) functional \cite{05YFW}
\begin{eqnarray}
&& {\cal F}(b)=-\sum_i  c_i b_i(x_i=1)-T {\cal S}(b),\label{BFE}\\
&& -{\cal S}(b)=\sum_j\sum_{x_{\mathcal S_j}} b_j(x_{\mathcal S_j})\log(b_j(x_{\mathcal S_j})) \nonumber\\
&& +\sum_i (1-q_i)\sum_{x_i} b_i(x_i)\log(b_i(x_i)),
\label{Entropy}
\end{eqnarray}
where $q_i=\sum_{j:A_{ji\neq 0}}1$,
subject to the following normalization and local consistency constraints:
\begin{eqnarray}
&&\sum_{x_{\mathcal S_j}:x_i} b_j \left( x_{\mathcal S_j} \right) = b_i(x_i),\quad\forall i\in \mathcal S_j \label{eq1-bplp}\\
&& b_j(x_{\mathcal S_j})\geq 0, \quad\sum_{x_{\mathcal S_j}} b_j (x_{\mathcal S_j}) = 1,\label{eq2-bplp}\\
&& b_j(x_{\mathcal S_j}) =0,\ \mbox{if}~\sum_{i \in \mathcal S_j} A_{ji}x_i > d_j,\ \forall j.
\label{eq3-bplp}
\end{eqnarray}
Note that we use $b_i(x_i=1)$ to mean $b_i(1)$.

It is known \cite{05YFW} that if BP converges, it finds a minimum (possibly local) of the BFE. Finding the global minimum of the BFE is desirable (as an approximation).  This task is reduced at $T=0$ to the following LP:
\begin{eqnarray}
\mbox{BPLP}:\ \min -\sum_i c_i b_i(x_i = 1),\ \mbox{s.t. ~(\ref{eq1-bplp}),~(\ref{eq2-bplp}),~(\ref{eq3-bplp})}.
\label{BPLP}
\end{eqnarray}

\vspace{-4pt}
\subsection{Illustrative Example: ILP and LP for Matching}\label{sec:ILP_and_LP_for_Matching}
\vspace{-2pt}
We illustrate the ILP formulation and transformation to a GM described in Section \ref{sec:ILP_as_MAP} on the weighted matching problem. Given an (undirected) graph $G=(V,E)$ with non-negative edge weights $ \{w_e:e\in E\}$, we seek to find the matching of largest weight, where a matching is a subset of edges such that each vertex is incident to at most one edge. The problem is described by the following ILP:
\begin{eqnarray}
\mbox{m-ILP:}\quad &&\max~ \sum_{e\in E} w_e x_e \label{m-ILP}\\
&&\mbox{s.t.}\qquad\sum_{e\in \delta(i)} x_e\leq 1,\quad\forall i\in V;\quad x_e\in \{0,1\}.
\nonumber
\end{eqnarray}
 where $\delta(i)=\{e=(i,j)\in E\}$ is the set of edges adjacent to vertex $i$.

The straightforward LP relaxation of m-ILP is formed by replacing $x_e \in \{0,1\}$ by $x_e \in [0,1]$. However, this LP is not tight in general - i.e. $\mbox{m-LP}\geq \mbox{m-ILP}$. The LP can be made tight, as famously shown by Edmonds \cite{Edmonds1965}, by adding a set of \emph{blossom} inequalities:
\begin{eqnarray}
\mbox{m-bl-LP}:\qquad&\max~ \sum_{e\in E} w_e x_e \label{m-LP}\\
\mbox{s.t.}\qquad&\sum_{e\in \delta(i)} x_e\leq 1,\quad\forall i\in V	\nonumber \\
\qquad\qquad&\sum_{e\in E(S)} x_e\leq \frac{|S|-1}2,\quad\forall S\in\mathcal S \nonumber \\
\qquad\qquad &x_e\in [0,1].\nonumber
\end{eqnarray}
where $E(S)=\{(i,j)\in E:i,j\in S\}$ is the set of edges with both ends in $S$ and $\mathcal S \subset 2^V$ is the set of all odd-sized sets of vertices in $G$. The blossom inequalities imply that an odd cycle of length $2l+1$ can have $l$ edges in a matching.

The weighted matching problem can be formulated as a MAP inference problem by associating a random variable with each edge $X=[X_e]\in \{0,1\}^{|E|}$ and constructing the following distribution:
\begin{eqnarray}
\Pr[X=x] ~\propto~ \prod_{e\in E}e^{w_ex_e} \prod_{i\in V} \psi_i(x_i) \prod_{S\in\mathcal S} \psi_S(x_S),
\label{GM-m}\\
\psi_i(x_i)=\begin{cases}
1, &\mbox{if}~\sum_{e\in \delta(i)} x_e \leq 1\\
0,&\mbox{otherwise}
\end{cases} \label{vertex}\\
\psi_S(x_S)=
\begin{cases}
1, &\mbox{if}~\sum_{e\in E(S)} x_e\leq \frac{|S|-1}2 \\
0, &\mbox{otherwise}
\end{cases}\label{blossom}.
\end{eqnarray}
where $\psi_i$ are functions defined over variables $x_i = \{x_e : e \in \delta(i)\}$ and $\psi_S$ are functions defined over $x_S = \{x_e : e \in E(s) \}$. It is easy to see that (\ref{GM-m}) is equivalent to
\begin{equation}
\Pr[X=x] ~\propto~\begin{cases} \exp \left( w(x) \right) &\mbox{if $x$ induces a matching in $G$}\\0&\mbox{otherwise}\end{cases}, \label{GM-m1}
\end{equation}
where $w(x):=\sum_{e\in E} w_e x_e$.

\section{Equivalence between LP and BPLP}\label{sec:main}
\vspace{-2pt}
Now we state the main result of the paper.
\begin{theorem}\label{thm1}
For any (fixed) $j$, consider the polytope
\begin{eqnarray}
P_j:\qquad\sum_{i \in \mathcal S_j} A_{ji}x_i\leq d_j,\quad
0\leq x_i\leq1\quad\forall i\in \mathcal S_j.\label{eq:cond1}
\end{eqnarray}
Then, the following properties hold:
\begin{itemize}
\item If $P_j$ has only 0-1 integral vertices (i.e., extreme points) for all $j$, then LP $\leq$ BPLP.
\item LP $\geq$ BPLP (without any conditions).
\end{itemize}
\end{theorem}
Theorem \ref{thm1} implies the following corollary.
\begin{corollary}\label{cor1}
If $A_{ji}\in\{-1,0,1\}$ for all $i,j$, then LP $=$ BPLP.
\end{corollary}
\vspace{-4pt}
\begin{proof}
Corollary \ref{cor1} is proved using Theorem \ref{thm1} and the fact that each vertex of a polytope can be expressed as the unique solution to a system of `face' linear equalities (see e.g. \cite{schrijver04}.
\end{proof}
\vspace{-4pt}
Note that the condition of Corollary \ref{cor1} holds for m-bl-LP. One also observes (arguing by contradiction) that the condition in Theorem \ref{thm1} for LP $\leq$ BPLP is necessary.
For example, suppose the number of rows of matrix $A$ is one, $S_1=\{1,\dots, n\}$ and the polytope $P_1$ has a fractional vertex $x=[x_i]$. Then there exists $c=[c_i]$ such that $x$ is the unique solution of LP (\ref{LP}). However, $[b_i(1)]=[x_i]$ cannot satisfy (\ref{eq1-bplp}), (\ref{eq2-bplp}) and (\ref{eq3-bplp}) for any factor $b_j(\cdot)$ because $x$ is a fractional vertex of $P_1$. Hence, LP $>$ BPLP.

\vspace{-4pt}

\subsection{Proof for LP $\leq$ BPLP}
\vspace{-2pt}
Here we prove that if $x=[x_i]$ satisfies the constraints of LP (\ref{LP}), then
there exists normalized beliefs $\{b_j\}$ such that $[b_i(1)]=[x_i]$, $[b_i(0)]=[1-x_i]$ and $\{b_i,b_j\}$ satisfies constraints of BPLP.
From the condition of Theorem \ref{thm1}, the polytope $P_j$ has only 0-1 integral vertices.
Then, according to the Carath\'{e}odory's theorem \cite{Cara_wiki}, any point $[x_i]$ in the polytope can be expressed as a convex combination of 0-1 vertices, where coefficients in the convex combination provide values of $\{b_j\}$, and the variables $b_i(0)$ and $b_i(1)$ in the description of the BPLP polytope, correspond to the variables $1-x_i$ and $x_i$ in the LP polytope, respectively.
This completes the proof of LP $\leq$ BPLP.

\vspace{-4pt}

\subsection{Proof for LP $\geq$ BPLP}
\vspace{-2pt}
Here we prove that if $\{b_i, b_j\}$ satisfies the constraints of BPLP, then $[x_i=b_i(1)]$ satisfies the constraints of LP as well. As mentioned above, $b_i(0)$ is redundant as $b_i(0)=1-b_i(1)$ within the BPLP polytope. From this, one derives
\begin{eqnarray*}
\sum_{i\in \mathcal S_j} A_{ji}x_i &=& \sum_{i\in \mathcal S_j} A_{ji} \sum_{x_{\mathcal S_j}:x_i=1} b_j \left( x_{\mathcal S_j}\right)\\
&=&
\sum_{i\in \mathcal S_j} A_{ji} \sum_{x_{\mathcal S_j}} x_i b_j\left( x_{\mathcal S_j}\right)\\
&=&\sum_{x_{\mathcal S_j}} \left(\sum_{i\in \mathcal S_j}A_{ji}x_i\right)b_j\left( x_{\mathcal S_j}\right)
\end{eqnarray*}
\begin{eqnarray*}
&\leq&\sum_{x_{\mathcal S_j}} d_j b_j\left( x_{\mathcal S_j}\right)\\
&=&d_j\sum_{x_{\mathcal S_j}} b_j\left( x_{\mathcal S_j}\right) = d_j, \\
\end{eqnarray*}
where (\ref{eq3-bplp}) was used at the inequality stage. This completes the proof of LP $\geq$ BPLP.

\section{Annealing BP for Solving LPs}\label{sec:annealbp}
\vspace{-2pt}
In this section, we propose the following annealing version of BP as an LP solver:
\begin{align*}
m_{ji}^{t+1}(x_i) & \leftarrow
m_{ji}^{t}(x_i)^{1-\alpha_t}
\sum_{x_{\mathcal S_j} \setminus x_i} \psi_j(x_{\mathcal S_j})^{\frac{\alpha_t}{T_t}} \prod_{k \in \mathcal S_j \setminus i} m_{kj}^t(x_k) ^{\alpha_t}\nonumber \\
m_{ij}^{t+1}(x_i) & \leftarrow m_{ij}^{t}(x_i)^{1-\alpha_t} \exp\left(\frac{\alpha_t c_i}{T_t} x_i \right) \prod_{k \in \mathcal E_i \setminus j} m_{ki}^t(x_i)^{\alpha_t} \nonumber\\
b_j^{t+1}(x_{\mathcal S_j}) & \propto  \psi_j(x_{\mathcal S_j})^{1/T_t} \prod_{k \in \mathcal S_j} m_{kj}^t(x_k) \nonumber \\
b_i^{t+1}(x_i) & \propto  \exp\left(\frac{c_i}{T_t} x_i \right) \prod_{k \in \mathcal E_i} m_{ki}^t(x_i), \nonumber
\end{align*}
where $\alpha_t\in(0,1]$, $T_t>0$, $m^{t}_{ij}, m^{t}_{ji}$ and $b_j^t, b_i^t$ are a `damping' parameter, a temperature parameter, messages and beliefs at the $t$-th iteration, respectively.
We have the following conjecture.
\begin{conjecture}
If LP $=$ BPLP and $m_{ij}^{0}=m_{ji}^{0}=1$ for all $i,j$, then there exists a scheme with annealing schedule $T_0\geq T_1\geq \dots$ with $\lim_{t\to\infty} T_t=0$ and damping schedule
$\alpha_0,\alpha_1,\dots$ such that $[b_i^t(1)]$ converges to the solution of LP.
\end{conjecture}

We now explain the rationale for the above conjecture. First, recall the following facts:
\begin{itemize}
\item If BP converges, it finds a (possibly local) minimum of the BFE function.
\item The BFE minimization is equal to BPLP at $T=0$.
\end{itemize}
The main difficulties in establishing the conjecture are (a) BP may not converge, and (b) BP may converge to a local (not global) minimum of the BFE functional. To overcome both issues, one can use a convex modification of the BFE function \cite{07WY}, and the known convergent variant to BP (providing sufficient damping), called CCCP, to find its minimum \cite{02Yuille}. We believe that an appropriate annealing scheme can fix the convergence issue and that the natural initialization $m_{ij}^{0}=m_{ji}^{0}=1$ can prevent annealing BP from converging to an undesirable local minimum of the BFE functional. Support for natural message initialization comes from \cite{08BSS,11SMW,10GSY,07SSW}, where for certain GMs the natural initial messages are needed to prevent BP from converging to an undesirable fixed point. We empirically verify this conjecture for matching GMs in the following section.

\section{Experiments with Matchings}\label{sec:exp}
\vspace{-2pt}

\begin{figure}
\begin{center}
\includegraphics[width=7.5cm]{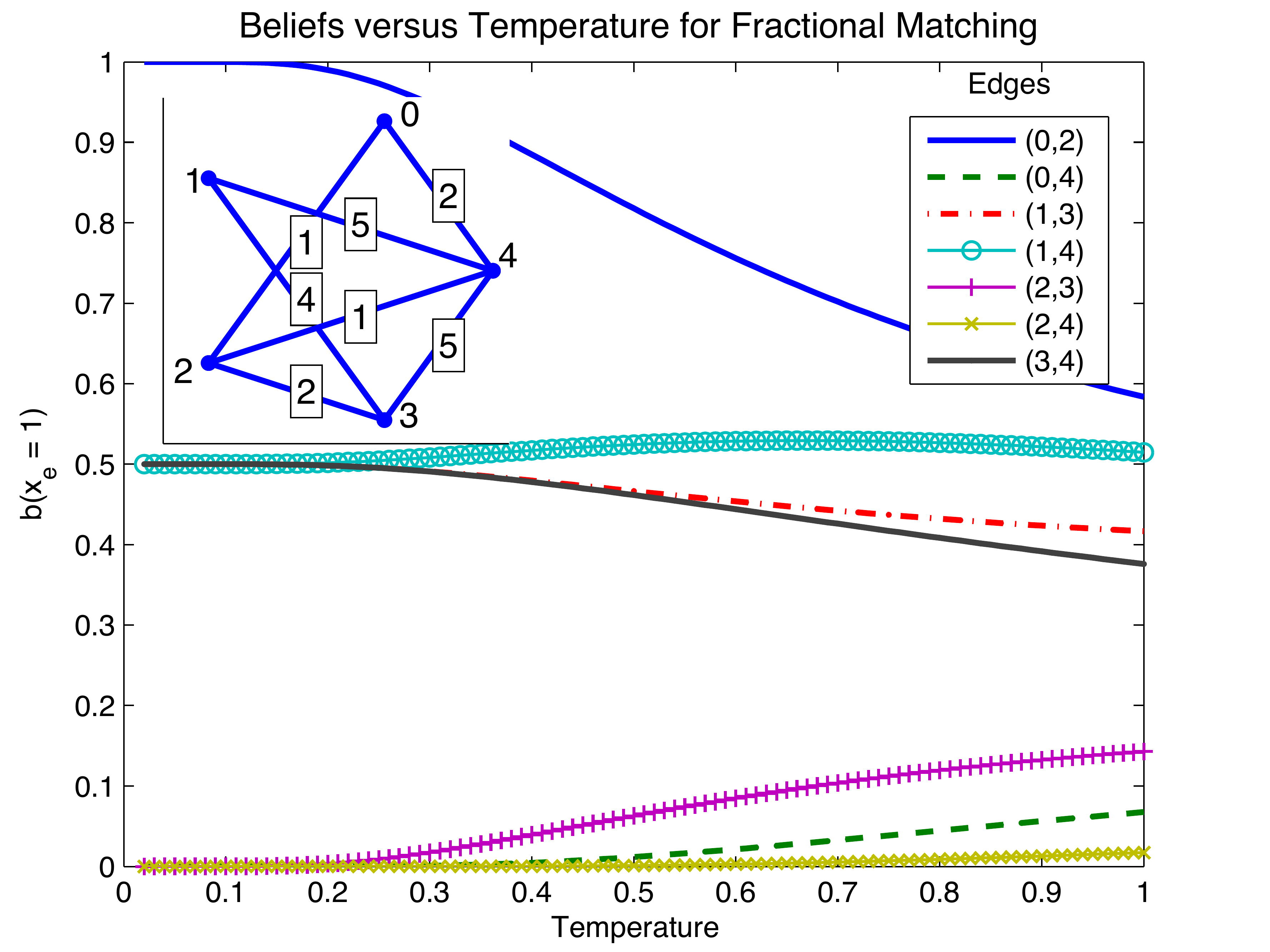}
\end{center}
\vspace{-6pt}
\caption{\small Convergence of edge beliefs found by annealing BP to a fractional LP solution with total weight $w(x) = 8$.}
\label{fig:Match5_fractional}
\vspace{-6pt}
\end{figure}

\begin{figure}
\begin{center}
\includegraphics[width=7.5cm]{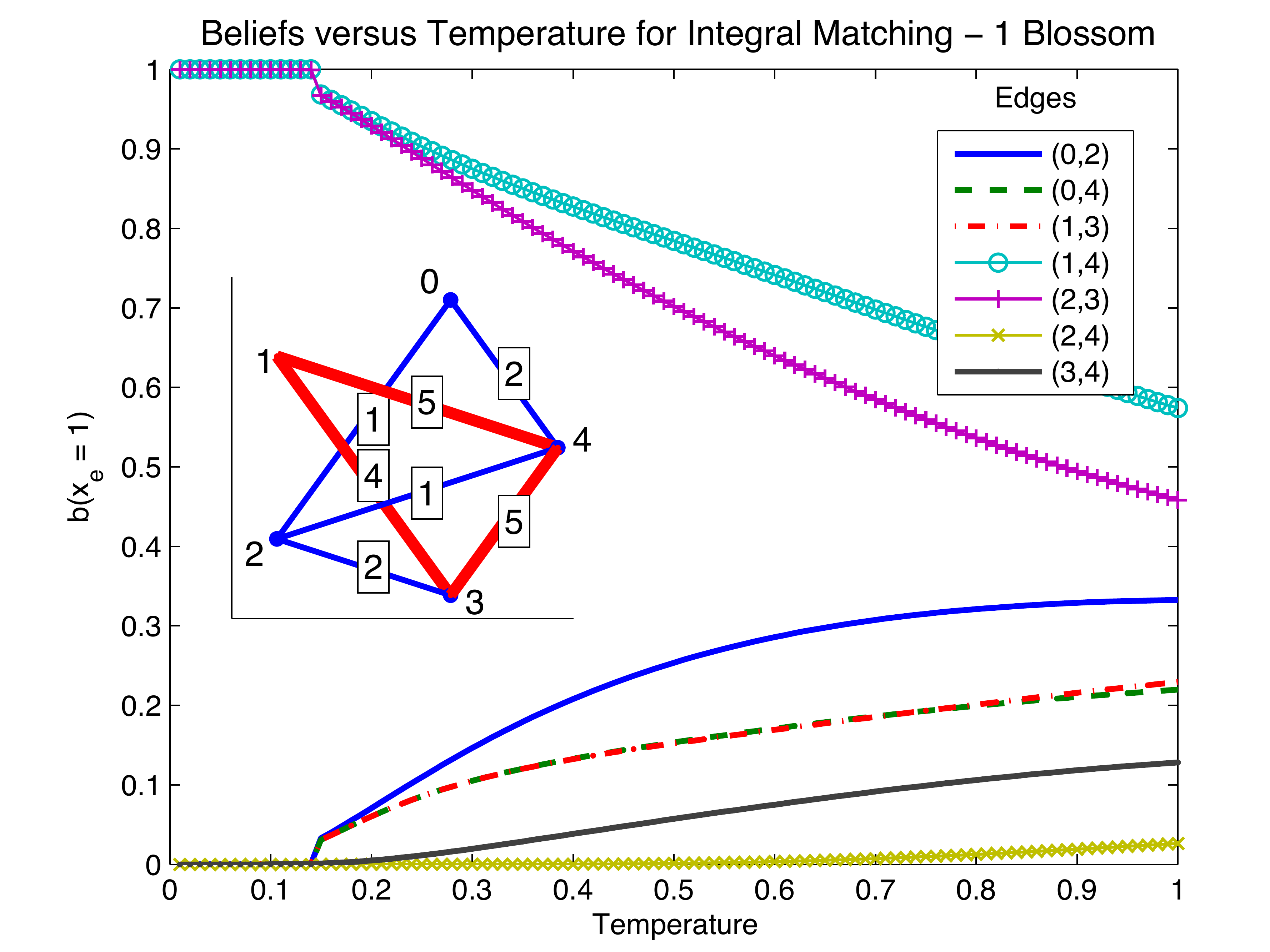}
\end{center}
\vspace{-4pt}
\caption{\small Convergence of edge beliefs found by annealing BP to the integral LP solution with total weight $w(x) = 7$.}
\label{fig:Match5_integral_blossom}
\vspace{-2pt}
\end{figure}

In this section, we demonstrate that annealing BP with sufficient damping can be used to reliably solve sequential LP relaxations to the weighted matching problem introduced in section \ref{sec:ILP_and_LP_for_Matching}. We note that our approach here differs from prior work on solving the weighted matching problem using BP because we consider the sum-product form of BP.  The work in \cite{08BSS,11SMW} demonstrated that max-product BP will converge to the MAP solution (and therefore find the maximum weight matching) if the relaxation to the matching ILP without blossoms is tight. However, when this LP is not tight, max-product will fail to converge. The connection between BPLP and LP made in the previous section, tells us that the MAP solution will correspond to the solution to the LP involving blossoms (i.e. m-bl-LP). We demonstrate that annealed sum-product BP can be used to solve m-bl-LP.

\begin{figure*}[t!]
\begin{center}
\includegraphics[width=5.8cm]{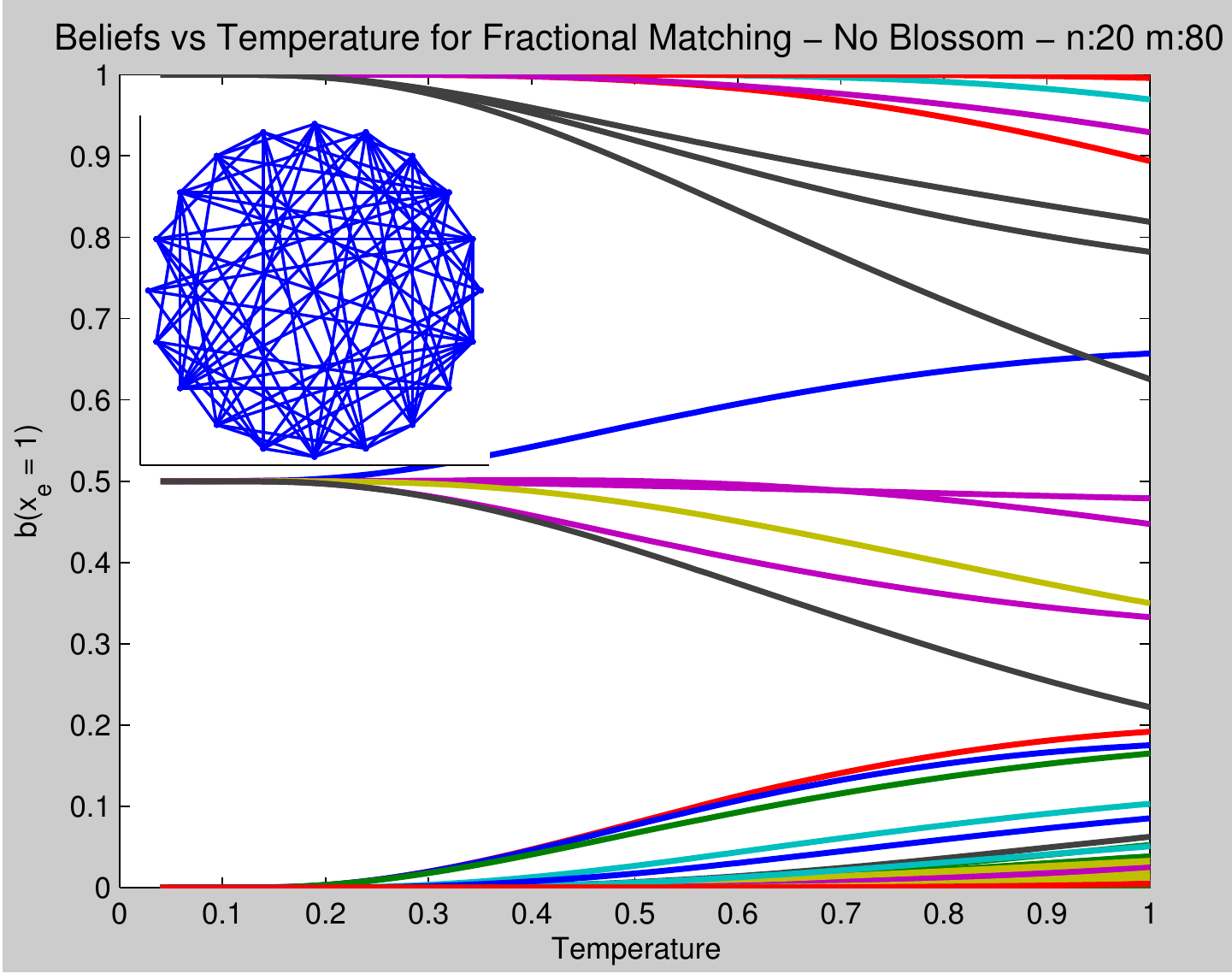}
\includegraphics[width=5.8cm]{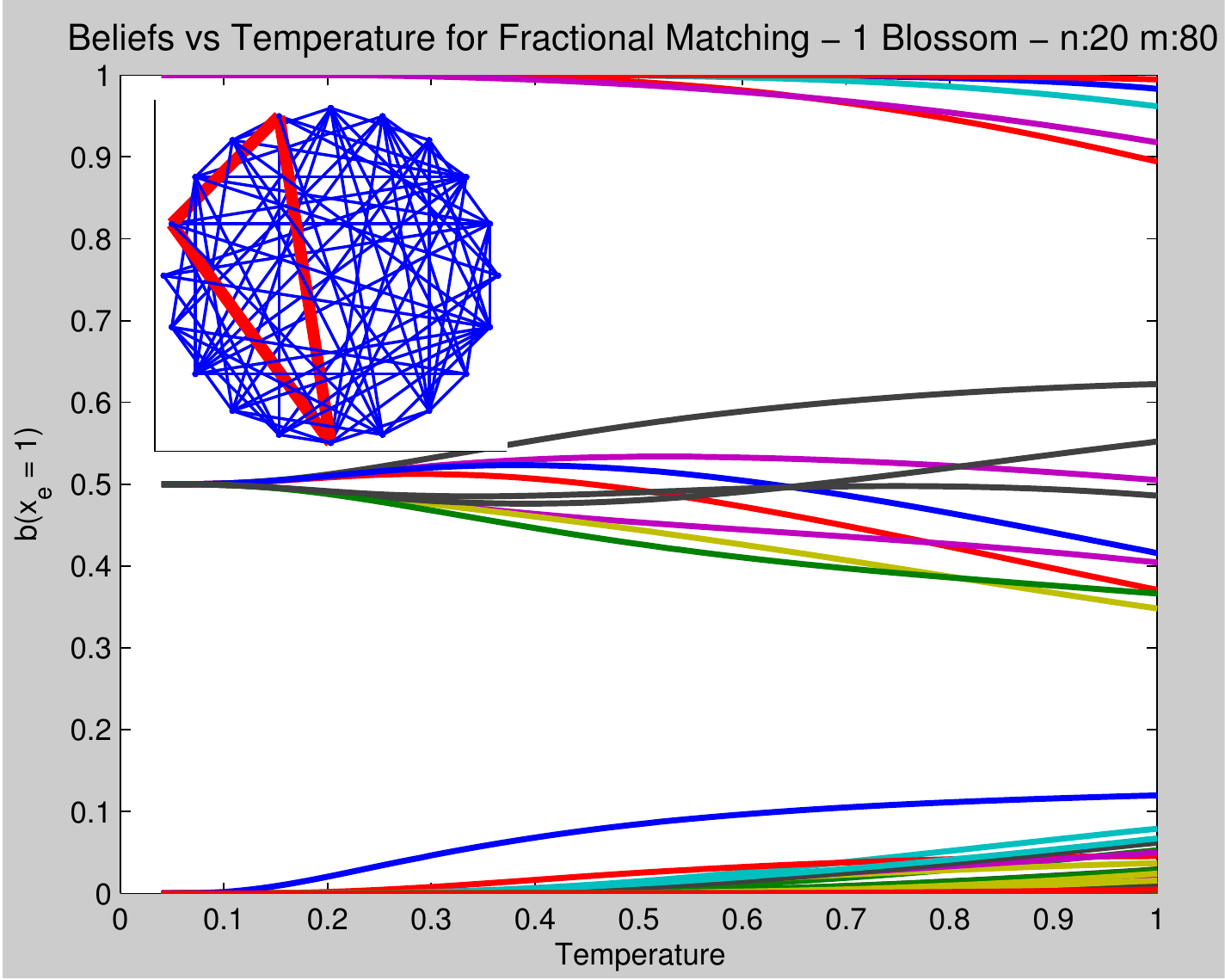}
\includegraphics[width=5.8cm]{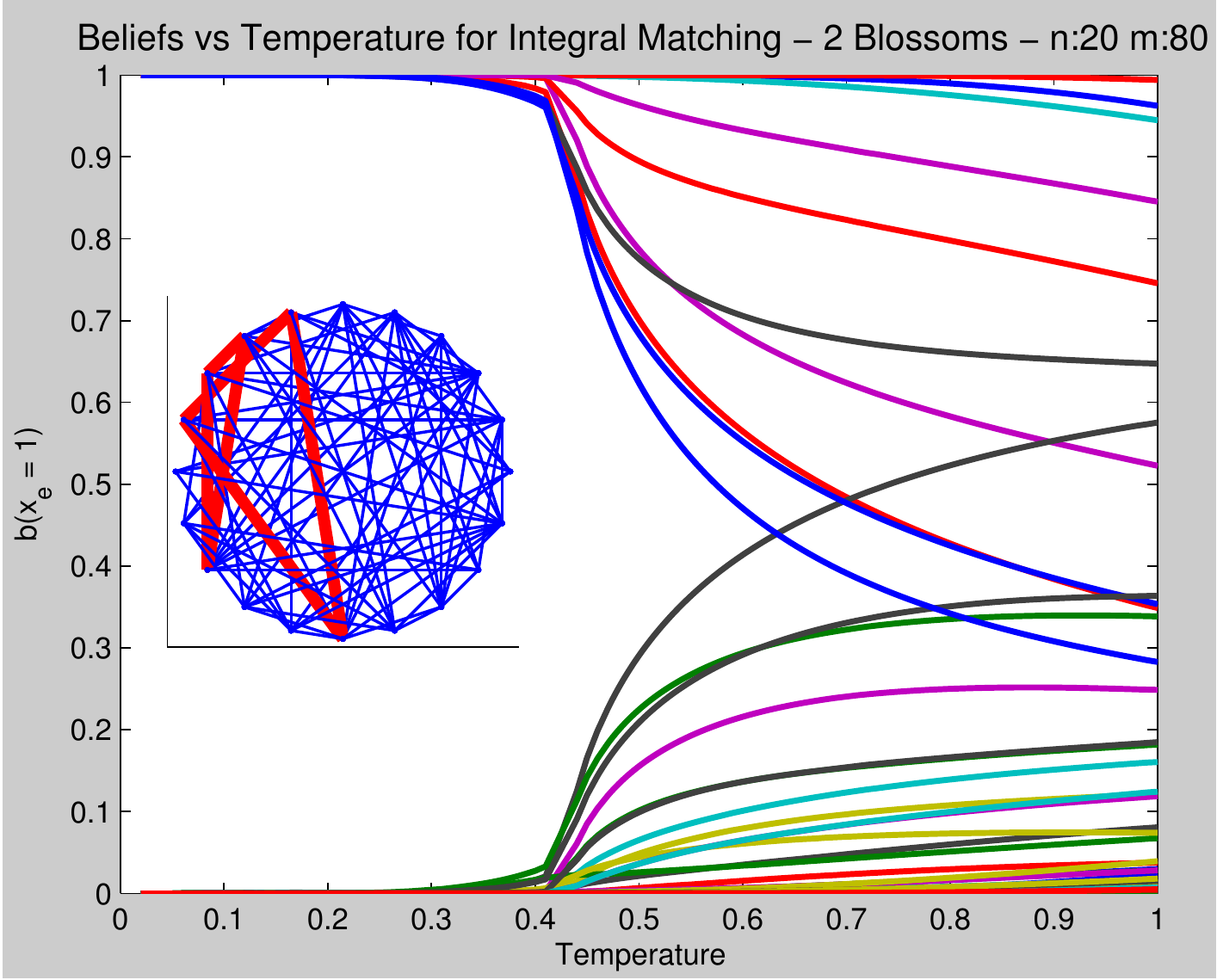}
\end{center}
\vspace{-6pt}
\caption{\small {\bf Left:} Relaxation of m-ILP without blossoms that is fractional. {\bf Middle:} Relaxation of m-ILP with a single blossom (shown in bold) that is still fractional. {\bf Right:} Relaxation of m-ILP with two blossoms (both in bold) that is tight (integral).}
\label{fig:Match20}
\vspace{-6pt}
\end{figure*}

Figures \ref{fig:Match5_fractional} and \ref{fig:Match5_integral_blossom} plot the edge beliefs $b_e(x_e = 1)$ found by sum-product BP on a particular weighted matching problem instance with $5$ vertices and $7$ edges. The edge weights for this instance are depicted in the inlay of each figure. In both figures, $T$ is annealed linearly from $1$ to $0.01$ over $100$ steps and BP is run for $20$ iterations at each temperature with $\alpha_t = \frac{1}{2}$ for all $t$. The results in Figure \ref{fig:Match5_fractional} are for a GM corresponding to a non-tight relaxation of m-ILP. Notice as the temperature is annealed that the beliefs converge to a fractional solution with total weight $w(x) = 8$. In this plot, we see that vertices $\{1,3,4\}$ constitute an odd set of vertices violating a blossom inequality. Adding an inequality for this blossom makes m-bl-LP tight. The GM used in Figure \ref{fig:Match5_integral_blossom} includes an additional factor enforcing the blossom inequality. The beliefs in the tight GM converge to an integral (and exact) solution with total weight $w(x) = 7$.

Figure \ref{fig:Match20} demonstrates how annealing sum-product BP can be used to solve a much larger weighted matching problem, with $20$ vertices and $80$ edges. We use the same annealing and damping scheme as in the previous experiments. The left-most figure plots edge beliefs as a function of temperature for the LP relaxation without blossoms. This relaxation is not tight, so several edge beliefs converge to $b(x_e = 1) = \frac{1}{2}$. In the center plot we have added a single blossom constraint that also yields a fractional solution. However, the blossom tightens the LP relaxation, reducing the total weight from $w(x) = 87.5$ to $w(x) = 86.5$. The right-most plot depicts a tight LP relaxation (i.e. m-bl-LP $=$ m-ILP). Notice that as temperature is annealed, all edge beliefs converge to either $0$ or $1$. In the exact matching $w(x)=86$.

\vspace{-4pt}
\bibliographystyle{IEEEtran/IEEEtran}
\bibliography{refs}
\end{document}